\documentclass{article}
\PassOptionsToPackage{numbers}{natbib}
\usepackage[final]{neurips_2025}
\usepackage[utf8]{inputenc}
\usepackage[T1]{fontenc}
\usepackage{hyperref}
\usepackage{url}
\usepackage{booktabs}
\usepackage{amsfonts}
\usepackage{nicefrac}
\usepackage{microtype}
\usepackage{graphicx}
\usepackage{multirow}

\title{ATANT: An Evaluation Framework for AI Continuity}

\author{
  Samuel Sameer Tanguturi \\
  Kenotic Labs \\
  \texttt{sam@kenoticlabs.com} \\
}

\begin{document}

\maketitle

\vspace{-0.5em}
\noindent\footnotesize
\textbf{Resources:}
Project: \url{https://kenoticlabs.com} \quad
Thesis: \url{https://kenoticlabs.com/thesis} \quad
Dataset: \url{https://huggingface.co/datasets/Kenotic-Labs/ATANTV1.0-corpus} \quad
Code: \url{https://github.com/Kenotic-Labs/ATANT}
\normalsize
\vspace{0.5em}

\begin{abstract}
We present ATANT (Automated Test for Acceptance of Narrative Truth), an open evaluation framework for measuring \textit{continuity} in AI systems: the ability to persist, update, disambiguate, and reconstruct meaningful context across time. While the AI industry has produced memory components (RAG pipelines, vector databases, long context windows, profile layers), no published framework formally defines or measures whether these components produce genuine continuity. We define continuity as a system property with 7 required properties, introduce a 10-checkpoint evaluation methodology that operates without an LLM in the evaluation loop, and present a narrative test corpus of 250 stories comprising 1,835 verification questions across 6 life domains. We evaluate a reference implementation across 5 test suite iterations, progressing from 58\% (legacy architecture) to 100\% in isolated mode (250 stories) and 100\% in 50-story cumulative mode, with 96\% at 250-story cumulative scale. The cumulative result is the primary measure: when 250 distinct life narratives coexist in the same database, the system must retrieve the correct fact for the correct context without cross-contamination. ATANT is system-agnostic, model-independent, and designed as a sequenced methodology for building and validating continuity systems. The framework specification, example stories, and evaluation protocol are available at \url{https://github.com/Kenotic-Labs/ATANT}. The full 250-story corpus will be released incrementally.
\end{abstract}

\section{Introduction}

Most AI systems today are session-based. A user provides input, the system responds, and the moment ends. Whatever persists is typically prompt context, conversation history, or retrieved notes. This is adequate for single-turn tasks but insufficient for any system that claims to maintain a meaningful relationship with a user over time.

Human interaction with AI extends beyond isolated prompts. Users have unfinished situations, changing states, recurring concerns, and ongoing commitments. AI systems that serve users across time need more than single-session intelligence. They need structured persistence.

The industry has produced partial solutions: long context windows keep recent material alive temporarily; RAG pipelines retrieve semantically similar text from storage; profile layers hold static preferences; vector databases store embeddings for similarity search. None of these, individually or combined, produce what we term \textbf{continuity}: the system property that determines what persists, in what form, what has changed, what still matters, and how to reconstruct it.

Despite growing recognition of this gap \citep{logan2026cma, natangelo2025nct, packer2023memgpt, chhikara2025mem0}, the field lacks three things:

\begin{enumerate}
    \item A formal definition of what continuity means as a system property.
    \item A set of testable requirements that any continuity system must satisfy.
    \item A benchmark that measures continuity rather than retrieval accuracy alone.
\end{enumerate}

ATANT addresses all three. Our contributions are:

\begin{itemize}
    \item A formal definition of \textbf{continuity} as an architectural layer, distinct from memory, retrieval, and context, with 7 required properties.
    \item A \textbf{model-independent evaluation methodology} with 10 checkpoints that tests the write path and read path of a continuity system without an LLM in the evaluation loop.
    \item A \textbf{narrative test corpus} of 250 stories (1,835 questions) across 6 life domains, designed for progressive evaluation from isolated correctness to disambiguation at scale.
    \item \textbf{4 compliance levels} (Core, Stress, Cumulative, Scale) that provide a sequenced roadmap for building and validating continuity systems.
    \item \textbf{Reference implementation results} across 5 test suite iterations, including honest reporting of failures, limitations, and the active frontier at 96\% cumulative-scale accuracy.
\end{itemize}

\section{Related Work}

\subsection{Memory Systems for AI}

Several systems have been proposed to give AI persistent memory. MemGPT \citep{packer2023memgpt} introduces an operating system metaphor with tiered memory (core memory always in context, archival memory stored separately). Mem0 \citep{chhikara2025mem0} provides a production-oriented memory layer that extracts facts, stores them, and manages updates across user, session, and agent scopes. A-MEM \citep{xu2025amem} proposes agentic memory with self-organizing capabilities for LLM agents.

These systems address components of persistence but do not formally define or test for continuity as a system property. A system can store and retrieve facts without satisfying disambiguation, reconstruction, or temporal ordering.

\subsection{Architectural Frameworks}

The Continuum Memory Architecture (CMA) \citep{logan2026cma} defines 6 behavioral properties for long-horizon LLM agents: persistence, selective retention, retrieval-driven mutation, associative routing, temporal continuity, and consolidation. This is the closest prior work to our formalization. However, CMA focuses on memory mechanisms rather than the higher-level logic of what persists and reconstructs, and does not provide an evaluation corpus.

The Narrative Continuity Test \citep{natangelo2025nct} proposes a conceptual framework for evaluating identity persistence across 5 dimensions. It provides theoretical grounding but no implementation or test corpus.

\subsection{Evaluation Benchmarks}

Existing benchmarks evaluate specific capabilities: MemoryBench \citep{ai2025memorybench} tests memory and continual learning; BEAM \citep{tavakoli2025beam} benchmarks long-term memory beyond a million tokens. These evaluate memory retrieval in isolation. ATANT differs in three ways: (1) it tests the full write-path-to-read-path pipeline, not retrieval alone; (2) it uses naturalistic multi-turn narratives, not synthetic fact pairs; (3) its cumulative mode tests disambiguation under memory load, a property no existing benchmark measures.

\subsection{Positioning}

ATANT is not a replacement for memory systems or retrieval benchmarks. It is a framework for evaluating whether the combination of components in a system produces \textit{continuity}, a higher-order property that emerges from correct persistence, update handling, temporal ordering, disambiguation, and reconstruction working together.

\section{Defining Continuity}
\label{sec:continuity}

\textbf{Continuity} is the system property that enables an AI to carry forward what still matters from prior interactions, update it when reality changes, and reconstruct useful context later in the appropriate form for the current situation.

\subsection{Continuity vs. Memory}

Memory stores the past. Continuity keeps the right parts of the past alive in the present. A database can store \texttt{(user, partner\_name, Mia)}. A continuity system can answer ``Tell me about my relationship'' with a reconstruction that includes Mia, where she lives, how the user feels about her, what changed last week, and what remains unresolved.

\subsection{Continuity vs. Retrieval}

Retrieval returns text similar to a query. Continuity reconstructs the current state of a situation, including what changed, what remains active, and what was superseded. The distinction is between similarity search and state reconstruction.

\subsection{The 7 Required Properties}

We define 7 properties that any system claiming continuity must satisfy. These properties were derived empirically by building a continuity system, running it against hundreds of real-world narratives, and identifying what breaks when each property is absent.

\begin{table}[h]
\centering
\caption{The 7 Required Properties of Continuity}
\label{tab:properties}
\begin{tabular}{@{}clp{8cm}@{}}
\toprule
\# & Property & Testable Requirement \\
\midrule
1 & Persistence Beyond Session & After ingesting facts, terminate and restart the process. All facts must be retrievable with identical accuracy. \\
2 & Update Handling & Ingest a fact, then an update. The system must return the current state and distinguish it from the previous state. \\
3 & Temporal Ordering & Ingest facts with temporal references. The system must return correctly resolved dates, sequencing, and status. \\
4 & Disambiguation & Ingest overlapping narratives. The system must return correct facts for the correct narrative without cross-contamination. \\
5 & Reconstruction & After multi-turn ingestion, the system must retrieve connected facts sufficient to reconstruct a situation, not isolated fragments. \\
6 & Model Independence & Ingest with one model (or none). Retrieve with another. Accuracy must not degrade. \\
7 & Operational Usefulness & The system must function across at least 2 distinct application domains without architectural modification to the continuity layer. \\
\bottomrule
\end{tabular}
\end{table}

\section{ATANT: Framework Design}
\label{sec:framework}

\subsection{Design Principles}

ATANT is built on 5 principles:

\textbf{Model Agnosticism.} ATANT evaluates the continuity layer, not the intelligence layer above it. No model (language, vision, or otherwise) is included in the evaluation loop. The continuity layer must be correct independent of whatever intelligence layer sits on top of it.

\textbf{Narrative Realism.} Test inputs are naturalistic multi-turn conversations. People do not speak in database format. A single utterance may contain identity, event, time, emotional state, entity, intent, and logistics simultaneously.

\textbf{Write Path + Read Path Verification.} Both directions are tested: did the system correctly store the facts (write path), and did it correctly retrieve and reconstruct the answer (read path)?

\textbf{Determinism.} Same input, same output, every time. No sampling variance.

\textbf{Progressive Difficulty (The Sequence).} Testing proceeds in phases: isolated $\rightarrow$ stress $\rightarrow$ cumulative $\rightarrow$ scale. Each phase tests a harder property. The sequence tells a team where their system stands and what to fix next.

\subsection{The 10-Checkpoint System}

ATANT defines 10 checkpoints grouped into write-path verification (CP1--CP4), read-path verification (CP5--CP8), and cross-cutting concerns (CP9--CP10).

\begin{table}[h]
\centering
\caption{ATANT Checkpoint System}
\label{tab:checkpoints}
\begin{tabular}{@{}cllp{5.5cm}@{}}
\toprule
CP & Path & Name & Pass Criteria \\
\midrule
1 & Write & Input Classification & Classification matches expected type \\
2 & Write & Fact Extraction \& Storage & All expected keywords in storage \\
3 & Write & Predictive Indexing & $\geq$1 predicted query per stored fact \\
4 & Write & Type Tagging & Type tags match expected categories \\
\midrule
5 & Read & Query Classification & Query type matches expected \\
6 & Read & Structural Matching & Correct fact in top-$k$ candidates \\
7 & Read & Convergence & Multi-fact candidate set returned \\
8 & Read & \textbf{Final Answer} & \textbf{All expected keywords in answer} \\
\midrule
9 & Cross & Temporal Reasoning & Temporal type and direction correct \\
10 & Cross & Contextual Adaptation & Emotion detection and direction correct \\
\bottomrule
\end{tabular}
\end{table}

\textbf{CP8 (Final Answer) is the definitive checkpoint.} All others are diagnostic; they identify where failures occur when CP8 fails.

\subsection{Compliance Levels}

ATANT defines 4 compliance levels with 3 scoring tiers (Gold: 100\%, Silver: 95--99\%, Bronze: 90--94\%):

\begin{table}[h]
\centering
\caption{ATANT Compliance Levels}
\label{tab:compliance}
\begin{tabular}{@{}llp{7cm}@{}}
\toprule
Level & Requirement & What It Proves \\
\midrule
ATANT-Core & 50 stories, isolated & Basic continuity across 6 life domains \\
ATANT-Stress & 250 stories, isolated & Continuity generalizes to novel patterns \\
ATANT-Cumulative & 50 stories, cumulative & Disambiguation when narratives coexist \\
ATANT-Scale & 250 stories, cumulative & Disambiguation at scale \\
\bottomrule
\end{tabular}
\end{table}

\section{Narrative Test Corpus}
\label{sec:corpus}

\subsection{Design Philosophy}

ATANT stories are narrative simulations: realistic multi-turn conversations spanning simulated hours, days, or weeks. They test continuity for \textit{human life}: personal, private, emotional, and ongoing. The domains were chosen because continuity is fundamentally about carrying a person's life forward, not task management.

Stories systematically include adversarial patterns: multi-fact utterances, shared-subject constructions (``My brother and I''), pronoun chains, temporal updates (``Actually, it moved to Thursday''), general knowledge traps (``What's the capital of France?''), emotional overlays, negation, and ambiguous predicates.

\subsection{Corpus Statistics}

\begin{table}[h]
\centering
\caption{ATANT Corpus Statistics}
\label{tab:corpus}
\begin{tabular}{@{}lrrr@{}}
\toprule
Phase & Stories & Questions & Purpose \\
\midrule
Phase 1 (Core) & 50 & 304 & 6 life domains \\
Phase 2 Round 2 & 50 & 367 & Generalization \\
Phase 2 Round 3 & 50 & 386 & Novel patterns \\
Phase 2 Round 4 & 50 & 380 & Edge cases \\
Phase 2 Round 5 & 50 & 398 & Adversarial \\
\midrule
\textbf{Total} & \textbf{250} & \textbf{1,835} & \\
\bottomrule
\end{tabular}
\end{table}

\subsection{Life Domain Coverage}

The corpus covers 6 domains: \textbf{Career} (interviews, promotions, layoffs), \textbf{Relationships} (partners, family, friendships), \textbf{Health} (medical, fitness, recovery), \textbf{Learning} (courses, certifications, study), \textbf{Daily Life} (routines, errands, hobbies), and \textbf{Life Events} (moves, births, deaths, marriages, milestones).

\subsection{Story Format}

Each story contains: metadata (ID, category, simulated duration), conversation batches with simulated timestamps, expected memory stores per batch, and verification questions with expected keywords. A question passes if all expected keywords appear in the system's retrieved answer (case-insensitive, substring-permissive).

\section{Experiments and Results}
\label{sec:results}

We evaluate the NURA Memory Pipeline (Kenotic Labs), the first system designed against the ATANT framework, across 5 test suite iterations. All evaluations are LLM-independent.

\subsection{Historical Progression}

\begin{table}[h]
\centering
\caption{Reference Implementation Progression}
\label{tab:progression}
\begin{tabular}{@{}llrrrl@{}}
\toprule
Suite & Date & Stories & Questions & CP8 Rate & Key Change \\
\midrule
Legacy & Jan 2026 & 50 & N/A & 58\% & Legacy scoring (with LLM) \\
Legacy+ & Feb 2026 & 50 & N/A & 72\% & Tuning gains \\
Legacy++ & Feb 2026 & 50 & N/A & 58\% & Regression from over-tuning \\
\midrule
1.0 & Mar 8 & 50 & 304/304 & 100\% & New architecture (no LLM) \\
1.1 & Mar 9 & 100 & 671/671 & 100\% & Stress round 2 \\
1.2 & Mar 10 & 150 & 1,057/1,057 & 100\% & Stress round 3 \\
2.0 & Mar 12 & 250 & 1,835/1,835 & 100\% & Full scale isolated \\
2.1 & Mar 14 & 50 (cumul.) & 304/304 & 100\% & Cumulative mode \\
\bottomrule
\end{tabular}
\end{table}

The legacy pipeline reached a ceiling at 58\% and exhibited regression under tuning pressure: optimizing for one narrative pattern broke retrieval for another. The redesigned architecture reached 100\% on 250 stories within 6 days (March 8--14), indicating that continuity is an architecture problem rather than a tuning problem.

\subsection{Cumulative Results}

Cumulative mode is where continuity is actually tested. When multiple life narratives coexist in the same database, the system must retrieve the correct fact for the correct context without cross-contamination.

\begin{table}[h]
\centering
\caption{Cumulative Mode Results}
\label{tab:cumulative}
\begin{tabular}{@{}lrrr@{}}
\toprule
Mode & Stories & Questions & CP8 Rate \\
\midrule
Isolated (250) & 250/250 & 1,835/1,835 & 100.0\% \\
Cumulative (50) & 50/50 & 304/304 & 100.0\% \\
Cumulative (250) & $\sim$210/250 & 1,761/1,835 & 96.0\% \\
\bottomrule
\end{tabular}
\end{table}

\subsection{Compliance Achieved}

\begin{table}[h]
\centering
\caption{Reference Implementation Compliance}
\label{tab:compliance_achieved}
\begin{tabular}{@{}lll@{}}
\toprule
Level & Status & Tier \\
\midrule
ATANT-Core & Pass & Gold \\
ATANT-Stress & Pass & Gold \\
ATANT-Cumulative & Pass & Gold \\
ATANT-Scale & In progress & Silver (96\%) \\
\bottomrule
\end{tabular}
\end{table}

\subsection{Failure Analysis}

\textbf{Suite 1.2 failures (12 stories, 15 questions):} The structural matcher failed on niche predicates, specifically questions about specialized hobbies (bonsai, falconry, cave exploration) where the predicate vocabulary was too specialized. The fix (Predicate Lexicon expansion) was architectural, not parameter tuning.

\textbf{250-story cumulative failures (74 questions):} Similarly-named predicates from different stories compete when 250 narratives coexist. The system must disambiguate by context, entity, and trace convergence. The 4\% gap represents the current frontier.

\textbf{CP4 failures (Type Tagging, 51.4\%):} Object type tagging fails on exotic domain-specific objects (``varroa mite,'' ``Paraloid B-72 adhesive''). These are diagnostic and do not affect CP8 accuracy.

\section{Discussion}
\label{sec:discussion}

\subsection{Continuity Is an Architecture Problem}

The legacy-to-current progression (58\% $\rightarrow$ 100\%) demonstrates that continuity cannot be achieved through scoring optimization alone. The legacy pipeline suffered from regressions under tuning pressure, a characteristic failure of systems without architectural continuity support. The breakthrough came from grammar-first classification, deterministic trace convergence, and structural matching. These were architectural decisions, not hyperparameter changes.

\subsection{Cumulative Mode Is the Real Test}

Isolated mode proves the pipeline works. Any reasonably engineered system should eventually pass. Cumulative mode tests what retrieval-based systems fundamentally struggle with: disambiguation under memory load. When 250 life narratives share storage, semantic similarity conflates similar-but-distinct events. This is the gap between retrieval and continuity.

\subsection{Model Agnosticism as Future-Proofing}

ATANT evaluates continuity without any model in the loop. This is a design principle, not a limitation. The intelligence layer is not static: today it is LLMs, tomorrow it may be vision models, world models, or embodied agents. A standard tied to any specific architecture dies when that architecture is superseded. The 7 properties and 10 checkpoints describe what continuity \textit{is}, not what today's AI happens to look like.

\subsection{Limitations}

\textbf{Keyword verification, not reconstruction quality.} CP8 checks whether expected keywords appear in the retrieved answer. A system could pass by returning relevant facts without coherence. Future versions should add reconstruction quality metrics.

\textbf{Single-author corpus.} All 250 stories were written by one author, limiting linguistic diversity and cultural representation.

\textbf{Single evaluated system.} Only one system has been evaluated against ATANT. The framework's value depends on independent systems being tested. We invite any team building AI continuity to run ATANT and publish results.

\textbf{English only.} The corpus does not test multilingual continuity.

\section{Conclusion}
\label{sec:conclusion}

We presented ATANT, an evaluation framework for AI continuity. ATANT provides a formal definition of continuity (7 properties), a model-independent evaluation methodology (10 checkpoints), a narrative test corpus (250 stories, 1,835 questions), and a sequenced compliance framework (4 levels). The reference implementation demonstrates that continuity is an architecture problem addressable through deterministic engineering rather than probabilistic tuning.

ATANT is designed to evolve. Version 1.0 defines the foundation. Future versions will add reconstruction quality metrics, multi-language narratives, proactive behavior testing, and community-contributed stories.

The framework specification and evaluation protocol are available at \url{https://github.com/Kenotic-Labs/ATANT}.

\section*{Acknowledgments}

ATANT was developed as part of the continuity layer architecture at Kenotic Labs.

\bibliographystyle{plainnat}
\bibliography{references}

\end{document}